\documentstyle[psfig,fullname]{article}

\title{Beyond Word $N$-Grams}
\author{
Fernando C. Pereira \qquad Yoram Singer \\
{\centering AT\&T Research}\\[1ex]
{\centering Naftali Tishby} \\
{\centering The Hebrew University}
}
\date{March 5, 1996}
\newcommand{\Seq}{w_1\cdots w_n}
\newcommand{\subseq}[3]{{#2}_{#1..#3}}
\newcommand{\prefix}[2]{{#1}_{..#2}}
\newcommand{\lmix}{\mbox{\it Lmix}}

\begin{document}
\maketitle
\begin{abstract}
We describe, analyze, and evaluate experimentally a new probabilistic
model for word-sequence prediction in natural language based on {\em
prediction suffix trees} (PSTs). By using efficient data structures,
we extend the notion of PST to unbounded vocabularies.  We also show
how to use a Bayesian approach based on recursive priors over all
possible PSTs to efficiently maintain tree mixtures.  These mixtures
have provably and practically better performance than almost any
single model.  We evaluate the model on several corpora. The low
perplexity achieved by relatively small PST mixture models suggests
that they may be an advantageous alternative, both theoretically and
practically, to the widely used $n$-gram models.
\end{abstract}

\section{Introduction}
Finite-state methods\index{finite-state methods} for statistical
prediction of word sequences in natural language have had an important
role in language processing research since the pioneering
investigations of Markov and Shannon \shortcite{Shannon}. It is clear
that natural texts are not Markov processes\index{Markov process} of
any finite order \cite{Good:langstat}, because of very long range
correlations between words in a text such as arise from subject
matter.  Nevertheless, low-order alphabetic $n$-gram models have been
used effectively in tasks such as statistical language identification,
spelling correction and handwriting transcription, and low-order word
$n$-gram models have been the tool of choice for language modeling in
speech recognition. The main problem with such fixed-order models is
that they cannot capture even relatively local dependencies that exceed
model order, for instance those created by long but frequent compound
names or technical terms. On the other hand, extending model order
uniformly to accommodate those longer dependencies is not practical,
since model size grows rapidly with model order.
 
Several methods have been proposed recently \cite{VMM,WST} to model
longer-range regularities over small alphabets while avoiding the size
explosion caused by model order. In those models, the length of
contexts used to predict particular symbols is adaptively extended as
long as the extension improves prediction above a given threshold.
The key ingredient of the model construction is the {\em prediction
suffix tree} (PST), whose nodes represent suffixes of past input and
specify a distribution over possible successors of the suffix. Ron
{\em et al.}~\shortcite{VMM} showed that under realistic conditions a
PST is equivalent to a Markov process of variable order and can be
represented efficiently by a probabilistic finite-state automaton.
In this paper we use PSTs as our starting point.

The problem of sequence prediction appears more difficult when the
sequence elements are {\em words} rather than characters from a small
fixed alphabet. The set of words is in principle unbounded, since in
natural language there is always a nonzero probability of encountering
a word never seen before.  One of the goals of this work is to
describe algorithmic and data-structure changes that support the
construction of PSTs over unbounded vocabularies.
We also extend PSTs with a wildcard symbol that can
match against any input word, thus allowing the model to capture
statistical dependencies between words separated by a fixed number of
irrelevant words.

The main contribution of this paper is to show how to build models
based on {\em mixtures} of PSTs. We use two results from machine
learning and information theory. The first is that a mixture of an
ensemble of experts (models) \index{mixture of experts (models)} with
suitably selected weights performs better than almost any individual
member of the ensemble \cite{Wegman,Experts}. The second result is
that within a Bayesian framework the sum over exponentially many trees
can be computed efficiently using the recursive structure of the tree,
as was recently shown by Willems {et al.} \shortcite{WST}. Our
experiments with algorithms based on those theoretical results show
that a PST mixture, which can be computed almost as easily as a single
PST, performs better than the maximum a posteriori (MAP) PST.

An important feature of PST mixtures is that they can be built by a fully
online (adaptive) algorithm. Specifically, updates to the model
structure and statistical quantities can be performed incrementally
during a single pass over the training data. For each new word,
frequency counts, mixture weights and likelihood values associated
with each relevant node are appropriately updated. There is not much
difference in learning performance between the online and batch modes,
as we will see. The online mode seems much more suitable for adaptive
language modeling\index{adaptive language modeling} over longer test
corpora, for instance in dictation or translation, while the batch
algorithm can be used in the traditional manner of $n$-gram models in
sentence recognition.

Two sets of priors are used in our Bayesian model. The first set
defines recursively the prior probability distribution over {\em all}
possible PSTs. The second set, which is especially delicate because
the set of possible words is not fixed, determines the probability of
observing a word for the first time in a given context. This includes
two possibilities: a completely new word, and a word previously
observed but not in the present context. We assign these priors using
a simplification of the Good-Turing method previously used in
compression algorithms.  It turns out that prediction performance is
not too sensitive to particular choices of priors.

Our successful application of mixture PSTs for word-sequence
prediction and modeling make them worth considering in applications
like speech recognition or machine translation if online adaptation of
an existing model to new material is required. Of course, these
techniques still fail to represent subtler aspects of syntactic and
semantic information. We plan to investigate how the present work may
be refined by taking advantage of distributional models of semantic
relations \cite{PTL}.

In the next sections we present PSTs and the data structure for the
word prediction problem. We then describe and shortly analyze the
learning algorithm. We also discuss several implementation issues.
We conclude with a evaluation of various aspects of the
model on several English corpora. 

\section{Prediction Suffix Trees over Unbounded Sets}
Our models operate on a set $U\subseteq \Sigma^\star$ of {\em
possible words} over an alphabet $\Sigma$. Since $U$ is intended to
represent the set of words of a natural language, we do not assume
that we know it in advance. A {\em prediction suffix tree\index{prediction
suffix tree (PST)}} (PST) $T$ over $U$ is a finite tree with nodes
labeled by distinct elements of $U^\star$ such that the root is
labeled by the empty sequence $\epsilon$, and if $s$ is a son of $s'$
and $s'$ is labeled by $a\in U^\star$ then $s$ is labeled by $wa$ for
some $w\in U$.  Therefore, in practice it is enough to associate each
non-root node with the first word in its label, and the full label of
any node can be reconstructed by following the path from the node to
the root. In what follows, we will often identify a PST node with its
label.

\begin{figure}[htb]
\raggedright{
\ \psfig{figure=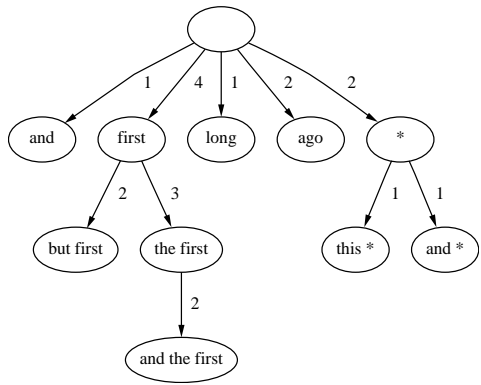,height=5.0cm}
\hspace*{0.30cm}{\parbox[b]{0.4\textwidth}{
\caption{A small example of a PST of words for language modeling. The 
numbers on the edges are the weights of the sub-trees starting at the 
pointed node. These weights are used for tracking a mixture of PSTs.
The special string {\tt *} represents a `wild-card' that can be matched 
with any observed word.
\label{PST:fig}}}}}
\end{figure}

Each PST node $s$ is has a corresponding {\em prediction function}
$\gamma_s : U^\prime \rightarrow [0,1]$, where $U^\prime \subset U
\cup \{\phi\}$. The symbol $\phi$ represents a {\em novel} event,
that is the occurrence of a word not seen before in the context
represented by $s$. The value of $\gamma_s$ is the {\em next-word
probability} function for the given context $s$. A PST $T$ can be
used to generate a stream of words, or to compute prefix
probabilities over a given stream. Given a prefix $w_1\cdots w_{k}$
generated so far, the context (node) used for prediction is found by
starting from the root of the tree and taking branches corresponding
to $w_{k}, w_{k-1}, \ldots$ until a leaf is reached or the next son
does not exist in the tree. Consider for example the PST shown in
Figure~\ref{PST:fig}, where some of the values of $\gamma_s$ are:
$$ \begin{array}{ll}
\gamma_{\tt `and \,
the \, first'}({\tt world}) = 0.1, &
\gamma_{\tt `and \, the \, first'}({\tt time}) = 0.6~ , \\
\gamma_{\tt `and \, the \, first'}({\tt boy}) = 0.2~, &
\gamma_{\tt `and \, the \, first'}(\phi) = 0.1 ~.
\end{array} $$
When observing the text `{\tt ... long ago and the first}', the
matching path from the root ends at the node {\tt `and the first'}.
Then we predict that the next word is {\tt time} with probability
$0.6$ and some other word not seen in this context with probability
$0.1$.  The prediction probability distribution $\gamma_s$ is
estimated from empirical counts. Therefore, at each node we keep a
data structure representing the number of times each word appeared in
the corresponding context.

The {\em wildcard symbol} `{\tt *}' allows a particular word position
to be ignored in prediction.  For example, the text `{\tt ... but this
was}' is matched by the node label `{\tt this *}', which ignores the
most recently read word `{\tt was}'.  Wildcards allow us to model
conditional dependencies of general form
$P(x_t|x_{t-i_1},x_{t-i_2},\ldots,x_{t-i_L})$ in which the indices
$i_1 < i_2 < \ldots < i_L$ are not necessarily consecutive.  Wildcards
provide a useful capability in language modeling since syntactic
structure may make a word depend less on the immediately preceding
words than on words further back.

One can easily verify that every standard $n$-gram\index{$n$-gram}
model can be represented by a PST, but the opposite is not true. A
trigram model, for instance, is a PST of depth two, where the leaves
are all the observed bigrams of words.  The prediction function at
each node is the trigram conditional probability of observing a word
given the two preceding words.

\section{The Learning Algorithm}
Within the framework of online learning\index{online algorithm}, it
can be proved \cite{Wegman,Experts} and demonstrated experimentally
that the performance of a weighted ensemble of models in which each
model is weighted according to its performance (the posterior
probability of the model), is not worse and generally much better
than any single model in the ensemble.  Although there might be
exponentially many different PSTs in the ensemble, it has been
recently shown \cite{WST} that a mixture of PSTs can be efficiently
represented for small alphabets.

We will use here Bayesian formalism to derive an {\em online}
learning procedure for mixtures of PSTs of words. The mixture
elements are drawn from some pre-specified set $\cal T$, which in our
case is typically the set of all PSTs with maximal depth $\leq D$ for
some suitably chosen $D$.

In what follows, we will consider a fixed input sequence $\Seq\cdots$.
To deal with boundary conditions, we will assume that the sequence is
padded on the left with enough ``start-of-sequence'' symbols. We will
denote by $\subseq{i}{w}{j}$ the input subsequence $w_{i+1}\cdots w_j$
and by $\prefix{w}{j}$ the prefix $w_1\cdots w_j$ (with appropriate
initial padding). By convention $\subseq{i}{w}{j} = \epsilon$
if $i \geq j$. For any PST $T$ and any sequence $s$, we will denote by
$s|T$ longest suffix of $s$ that has a corresponding node in $T$, and,
through our identification of PST nodes and sequences, the node itself.
Then $T$'s likelihood (or evidence) $P(\prefix{w}{n}|T)$ after observing
$\Seq$ is
\[
P(\prefix{w}{n}|T) = \prod_{i=1}^n \gamma_{\prefix{w}{i-1}|T}(w_i) \; .
\]
The probability of the next word given the past $n$ observations is 
then:
\begin{eqnarray}
P(w_{n+1}|\prefix{w}{n}) & = &
    {P(\prefix{w}{n+1}) \over P(\prefix{w}{n}) } \label{Bayes:eqn}
 \\
& = &
{{\sum_{T \in {\cal T}} {{P_0(T) P(\prefix{w}{n}w_{n+1}|T)}}} \over
{\sum_{T \in {\cal T}} P_0(T) P(\prefix{w}{n}|T)}} \; ,\nonumber
\end{eqnarray}
where $P_0(T)$ is the prior probability of the PST $T$.

A na\"{\i}ve computation of (\ref{Bayes:eqn}) would be infeasible,
because of the size of $\cal T$. Instead, we use a recursive method in
which the relevant quantities for a PST mixture are computed
efficiently from related quantities for sub-PSTs. In particular, the
PST prior $P_0(T)$ is defined as follows. A node $s$ has a probability
$\alpha_s$ of being a leaf and a probability $1 -\alpha_s$ of being an
internal node.  In the latter case, its sons are either a single
wildcard, with probability $\beta_s$, or actual words with probability
$1 - \beta_s$.  To keep the derivation simple, we assume here that the
probabilities $\alpha_s$ are independent of $s$ and that there are no
wildcards, that is, $\beta_s = 0 , \alpha_s = \alpha$ for all
$s$. Context-dependent priors and trees with wildcards can be obtained
by a simple extension of the present derivation. We also assume
that all the trees have maximal depth $D$. Then $P_0(T) = \alpha^{n_1}
(1 - \alpha)^{n_2}$, where $n_1$ is the number of leaves of $T$
of depth less than $D$ and $n_2$ is the number of internal nodes of $T$.

To evaluate the likelihood of the whole mixture we build a tree of
maximal depth $D$ containing all observation-sequence suffixes of
length up to $D$. The tree
built after observing $\Seq$ contains a node for each subsequence
$\subseq{i-k}{w}{i}$ for $0 \le k \le D$ and $1 \le i \le n$.
For each such node we keep two variables.\footnote{In practice, we keep
only a ratio related to the two variables, as explained in detail in
the next section.} The first, $L_n(s)$, accumulates the likelihood
the node would have if it were a leaf. That is, $L_n(s)$ is the
product of the predictions of the node on all the observation-sequence
suffixes that ended at that node:
\[
\begin{array}{rcl}
L_n(s) & = &
\prod_{\left\{i | \subseq{i-|s|}{w}{i} = s, \, 0 \le i < n \right\}}
	P(w_{i+1}|s) \\
& = &
\prod_{\left\{i | \subseq{i-|s|}{w}{i} = s, \, 0 \le i < n \right\}}
	\gamma_s(w_{i+1})
\end{array} \;\; .
\]
For each new observed word $w_n$, the likelihood values $L_{n+1}(s)$ are
derived from their previous values $L_{n}(s)$. Clearly, only the
nodes labeled by $\subseq{n-k}{w}{n}, \, 0 \le k \le D$ will need
likelihood updates. For those nodes, the update is simply
multiplication by the node's prediction for $w_{n+1}$, while for the rest
of the nodes the likelihood values do not change:
\begin{equation}
L_{n+1}(s) = \left\{
\begin{array}{ll}
L_{n}(s) \, \gamma_s(w_{n+1}) & s = \subseq{n-k}{w}{n} , \; 0 \le k \le D \\
L_{n}(s) & \mbox{otherwise}
\end{array} \right. \;\; .
\label{pred_update:eqn}
\end{equation}

The second variable, denoted by $\lmix_n(s)$, is the likelihood of
the mixture of {\em all} possible trees that have a subtree rooted at $s$
on the observed  suffixes (all observations that reached $s$). $\lmix_n(s)$ 
is calculated recursively as follows:
\begin{equation}
\label{mix_recur:eqn}
\lmix_{n}(s) = \alpha L_n(s) \;\;\; + \;\;\;
(1 - \alpha) \prod_{u \in U} \lmix_n(u s) \, ,
\end{equation}
The recursive computation of the mixture likelihood terminates at the leaves:
\[
\lmix_n(s) = L_n(s) \mbox{ if } |s| = D\, .
\]
The mixture likelihood values are updated as follows:
\begin{equation}
\lmix_{n+1}(s)\! = \!\left\{\!\!\!\!
\begin{array}{l@{\hspace{1ex}}l}
L_{n+1}(s) &  s=\subseq{n-D}{w}{n} \\
(1\! -\! \alpha) \prod_{u \in U} \lmix_{n+1}(u s)\! +\! \alpha L_{n+1}(s) &
s=\subseq{n-k}{w}{n},k < D \\
\lmix_{n}(s) & \mbox{otherwise} \;\; .
\end{array} \right.
\label{mix_update:eqn}
\end{equation}
At first sight it would appear that the update of $\lmix_{n+1}$ would require
contributions from an arbitrarily large subtree, since $U$ may be
arbitrarily large. However,
only the subtree rooted at $(w_{n-|s|}\,s)$ is actually affected by
the update. Thus the following simplification holds:
\[
\prod_{u \in U} \lmix_{n+1}(u s) \, = \, \lmix_{n+1}(w_{n-|s|}\,s) \;\; \cdot
\prod_{u \in U, u \neq w_{n-|s|}} \lmix_{n+1}(u s) \; .
\]

Note that $\lmix_n(s)$ is the likelihood of the weighted mixture of
trees rooted at $s$ on {\em all} past $n$ observations, where each tree
in the mixture is weighted with the appropriate  prior. Therefore
\begin{equation}
\lmix_n(\epsilon) = 
{{\sum_{T \in {\cal T}} {{P_0(T) P(\prefix{w}{n}|T)}}}} \;\; ,
\label{mix:eqn}
\end{equation}
where ${\cal T}$ is the set of trees of maximal depth $D$ and $\epsilon$
is the null context (the root node). Combining Equations (\ref{Bayes:eqn})
and (\ref{mix:eqn}), we see that the prediction of the whole mixture for
next word $w_{n+1}$ is the ratio of the likelihood values
$\lmix_{n+1}(\epsilon)$ and $\lmix_{n}(\epsilon)$ at the root node:
\[
P(w_{n+1}|\prefix{w}{n}) = \lmix_{n+1}(\epsilon) / \lmix_{n}(\epsilon) \; .
\]

A given observation sequence matches a unique path from the root to a
leaf. Therefore the time for the above computation is linear in the
maximal tree depth $D$. After predicting the next word
the counts are updated simply by increasing by one the count of the
word, if the word already exists, or by inserting a new entry for the
new word with initial count set to one. Our learning algorithm has,
however, the advantages of not being limited to a constant context
length (by setting $D$ to be arbitrarily large) and of being able to
perform online adaptation. Moreover, the interpolation weights
between the different prediction contexts are automatically
determined by the performance of each model on past observations.

In summary, for each observed word we follow a path from the root of
the tree corresponding to the previous words until a longest context
(maximal depth) is reached.  We may need to add new nodes, with new
entries in the data structure, for the first appearance of a word in a
given context.  The likelihood values of the mixture of subtrees
(Equation \ref{mix_update:eqn}) are returned from each level of that
recursion up to the root node. The probability of the next word is
then the ratio of two consecutive likelihood values returned at the root.

For prediction without adaptation, the same method is applied except
that nodes are not added and counts are not updated. If the prior
probability of the wildcard, $\beta$, is positive, then at each level
the recursion splits, with one path continuing through the node
labeled with the wildcard and the other through the node corresponding
to the proper suffix of the observation. Thus, the update or
prediction time is in that case $O(2^D)$. However, judicious use of
pruning can make the effective depth of the tree fairly small, making
update and prediction times linear in the text length.

It remains to describe how the probabilities, $P(w|s) = \gamma_s(w)$
are estimated from empirical counts. This problem has been studied for
more than thirty years and so far the most common techniques are based
on variants of the Good-Turing\index{Good-Turing} (GT) method
\cite{GT1,GT2}.  Here we give a description of the estimation method
that we implemented and evaluated.  We are currently developing an
alternative approach for cases when there is a known (arbitrarily
large) bound on the maximal size of the vocabulary $U$.  After
observing a certain training text, let $c_s(w)$ be the number of
occurrences of word $w$ in context $s$, $V_s$ be the set of words with
$c_s(w) > 0$, $r_s(n)$ be the number of distinct words observed
exactly $n$ times in context $s$, $r_s=\sum_{i}r_s(i)$ be the number of
distinct words observed following $s$, and $c_s = \sum_{w} c_s(w)$ the
total number of occurrences of $s$.  We seek estimates of
$\gamma_s(w)$ for $w\in V_s$ and of the probability $\gamma_s(\phi)$
observing after $s$ a word not in $V_s$ when scanning new text. The GT
method sets $\gamma_s(\phi) = {r_s(1) \over c_s}$.  This method has been
given several justifications, such as a Poisson assumption on the
appearance of new words \cite{Fisher}. It is, however, difficult to
analyze and requires keeping track of the rank of each word. Our
online learning method and data structures favor instead any method
that is based only on word counts. In source coding\index{source
coding} it is common to set the probability of a novel event in
context $s$ \index{source coding!novel event} to
$\gamma_s(\phi) = {r_s \over {n_s + r_s}}$, and allocate the remaining
probability mass to words in $V_s$ according to
$\gamma_s(w) = {c_s(w) \over {c_s + r_s}}$. Witten and Bell \shortcite{WB}
report that this method performs comparably to GT but is simpler because
it needs to keep track of the number of distinct words and their counts only.

We also need to distinguish between occurrences of completely new
words, that never occurred before in any context, and words that have
occurred before in other contexts but not the present one. A coding
interpretation helps us understand the issue.

Suppose text is being compressed according to the word probability
estimates given by a PST.  Whenever a completely new word is to be
encoded, we need to signal first that a new word occurred and then
encode the identity of the word, for instance with a lower level coder
based on a PST over the base alphabet\index{source coding!unbounded
alphabet} $\Sigma$ in which the words in $U$ are written. For a word
that has been observed before but not in the current context, it is
only necessary to code the identity of the word by referring to a
shorter context in which the word has already been observed (obviously
such a shortest context always exists) multiplying the probability of
the word in the shorter context by the probability that the word is
new in the longer context. This product is the probability of the
word in the next context.

In the case of a completely new word $w_n$, the probability of a
novel event must be multiplied by $P_0(w_n)$ corresponding to the
probability of $w_n$ according to the model in the lower-level
coder. The assumption that an independent lower-level coder is used
for completely new words implies that $P_0(w_n)$ is independent of the
context. We can thus factor $P_0(w_n)$ from $\lmix_m$ for $m \ge
n$. In particular, $P_0(w_n)$ will cancel out when calculating the
probability of the following word $w_{n+1}$:
\[P(w_{n+1}|\prefix{w}{n}) =
{{\lmix_{n+1}(\epsilon) \times P_0(w_n)} \over {\lmix_{n}(\epsilon)
\times P_0(w_n)}} =
{{\lmix_{n+1}(\epsilon)} \over {\lmix_{n}(\epsilon)}} \; .
\]
Therefore, for prediction we do not in fact need to use the
lower-level coder or estimate $P_0(w_n)$.

\section{Efficient Implementation of PSTs of Words}
We now describe refinements of the data structure and algorithm
described in the previous section to support very large vocabularies
and large training sets. The likelihood values $\lmix_n(s)$ and
$L_n(s)$ decrease exponentially with $n$, causing numerical problems
even if logarithmic representations are used. Moreover, we are only
interested in the predictions of the mixtures, not in the actual
likelihoods, which are only used to weigh the predictions of
different nodes. Let $\tilde{\gamma}_s(w_n)$ be the prediction of
the weighted mixture of all subtrees rooted below $s$ (including $s$
itself) for $w_n$. By following the derivation presented in the
preceding section it can be verified that
\[
\tilde{\gamma}_s(w_{n+1}) =
	q_n(s) \gamma_s(w_{n+1}) \; + \;
	(1 - q_n(s)) \tilde{\gamma}_{(w_{n-|s|}\,s)}(w_{n+1}) \;\; ,
\]
where
\[
\begin{array}{rcl}
q_n(s) & = & \alpha L_n(s) /
	\left(\alpha L_n(s) + (1 - \alpha) \prod_{u \in U} \lmix_n(u s)\right) \\
	& = &
		{1 / \left(1 + {(1 - \alpha) \prod_{u \in U} \lmix_n(u s) \over
		\alpha L_n(s)}\right)} \;\; .
\end{array}
\]
Define
\[
R_n(s) = \log\left(
	{\alpha L_n(s) \over (1 - \alpha) \prod_{u \in U} \lmix_n(u s)}\right) \;\; .
\]
Setting $R_0(s) = \log(\alpha / (1 - \alpha))$ for all $s$, 
$R_n(s)$ is updated as follows:
\[
R_{n+1}(s) = R_n(s) + \log\left(\gamma_s(w_{n+1})\right) -
	\log\left(\tilde{\gamma}_{(w_{n-|s|}\,s)}(w_{n+1})\right) \;\; ,
\]
and $q_n(s) = 1 / (1 + e^{-R_{n}(s)})$.
Thus, the probability of $w_{n+1}$ is propagated along the path
corresponding to suffixes of the observation sequence towards the
root as follows
\[
\tilde{\gamma}_s(w_{n+1}) = \left\{
\begin{array}{ll}
\gamma_s(w_{n+1}) & s=\subseq{n-D}{w}{n} \\
q_n(s) \gamma_s(w_{n+1}) + & \\
\;\;\;\; (1 - q_n(s)) \tilde{\gamma}_{(w_{n-|s|} s)}(w_{n+1})
& s=\subseq{n-k}{w}{n}, k < D
\end{array} \right.
\]

Finally, the prediction of the full mixture of PSTs for $w_n$ is simply
given by $\tilde{\gamma}_{\epsilon}(w_n)$.

For each node, we need frequency counts for the corresponding context
and for each word that occurred in that context. However, we can avoid
storing the word-in-context counts explicitly. To find how many times
word $w$ occurred in context $s=\subseq{n-k}{w}{n}$ we search the tree
for $s'=\subseq{n-k}{w}{n}w$. The number of times this node was
visited is also the number of occurrences of word $w$ in context
$s$. Therefore, each node is both used for predicting the next word
using a mixture of contexts of different lengths and for tracking the
number of times a given context has occurred. For a PST mixture of
bounded depth $D$, we thus need to maintain a tree of depth $D+1$. The
leaves of this tree are used for storing frequency counts for word
sequences of length $D+1$, while the internal nodes have the two roles
described above. In summary, for each node $s$ we maintain the count
$c_s$ of how many times the corresponding context has been
observed. In addition, for internal nodes we keep also the node's
log-likelihood ratio $R_n(s)$ described earlier, the number $r_s$ of
distinct words seen in context $s$ (the species count for the
context), and a list of the node's sons, that is, of contexts longer
by one word. When observing a word $w_{n}$, the online algorithm
performs two traversals of the tree, one to update the counts for
the nodes $\subseq{n-k}{w}{n}$, $0 \le k \le D+1$, and the
other to retrieve and update the log-likelihood ratio $R_n(s)$ and the
species count $r(s)$ for $s=\subseq{n-1-k}{w}{n-1}$, $0 \le k \le D$.

Natural language is often bursty \cite{CG96}, that is, rare
or new words may appear and be used relatively frequently for some
stretch of text only to drop to a much lower frequency of use for the
rest of the corpus.  Thus, a PST being build online may only need to
store information about those words for a short period.  It may then
be advantageous to prune PST nodes and remove small counts
corresponding to rarely used words. Pruning is performed by removing
all nodes from the suffix tree whose counts are below a threshold,
after each batch of $K$ observations. We used a pruning frequency $K$
of $1000000$ and a pruning threshold of 2 in some of our experiments.
\begin{figure}[htb]
{\small
\centerline{\fbox{\begin{minipage}{28pc}
every year public sentiment for conserving our rich natural
heritage is growing but that heritage is shrinking even faster no
joyride much of its contract if the present session of the cab driver
in the early phases conspiracy but lacking money from commercial
sponsors the stations have had to reduce its vacationing
\end{minipage}}}}
\caption{Text created by a random walk over a PST trained on the Brown corpus.}
\label{walk:fig}
\end{figure}

\begin{table}[th]
\label{on_line:tab}
\begin{center}
\begin{tabular}{|c|c|c|c|c|c|} \hline
{\em Text} & {\em Maximal} & {\em Number of} &
{\em Perplexity} & {\em Perplexity} & {\em Perplexity} \\
& {\em Depth} & {\em Nodes} &
$(\alpha = 0.5)$ & $(\alpha = 0.999)$ & $(\alpha = 0.001)$ \\ \hline

Bible      & 0 &      1 & 282.1 & 282.1 & 282.1 \\
(Gutenberg & 1 &   7573 &  84.6 &  84.6 &  84.6 \\
Project)   & 2 &  76688 &  55.9 &  58.2 &  55.5 \\
           & 3 & 243899 &  42.9 &  50.9 &  42.5 \\
           & 4 & 477384 &  37.8 &  49.8 &  37.5 \\
           & 5 & 743830 &  36.5 &  49.6 &  35.6 \\ \hline

Paradise Lost & 0 &      1 & 423.0 & 423.0 & 423.0 \\
by            & 1 &   8754 & 348.7 & 348.7 & 348.7 \\
John Milton   & 2 &  59137 & 251.1 & 289.7 & 243.9 \\
              & 3 & 128172 & 221.2 & 285.3 & 206.4 \\
              & 4 & 199629 & 212.5 & 275.2 & 202.1 \\
              & 5 & 271359 & 209.3 & 265.6 & 201.6 \\ \hline

Brown  & 0 &      1 & 452.8 & 452.8 & 452.8 \\
Corpus & 1 &  12647 & 276.5 & 276.5 & 276.5 \\
       & 2 &  76957 & 202.9 & 232.6 & 197.1 \\
       & 3 & 169172 & 165.8 & 224.0 & 165.6 \\
       & 4 & 267544 & 160.5 & 223.9 & 159.7 \\
       & 5 & 367096 & 158.7 & 223.8 & 158.7 \\ \hline
\end{tabular}
\caption{The perplexity of PSTs for the online mode.}
\end{center}
\end{table}
\begin{table}[th]
\begin{center}
\begin{tabular}{|c|c|c|c|} \hline
{\em Text} & {\em Maximal Depth} &
{\em Perplexity} ($\alpha = 0.5$) & {\em Perplexity} (MAP Model) \\ \hline
Bible      & 0 & 411.3 & 411.3 \\
(Gutenberg & 1 & 172.9 & 172.9 \\
Project)   & 2 & 149.8 & 150.8 \\
           & 3 & 141.2 & 143.7 \\
           & 4 & 139.4 & 142.9 \\
           & 5 & 139.0 & 142.7 \\ \hline

Paradise Lost & 0 & 861.1 & 861.1 \\
by            & 1 & 752.8 & 752.8 \\
John Milton   & 2 & 740.3 & 746.9 \\
              & 3 & 739.3 & 747.7 \\
              & 4 & 739.3 & 747.6 \\
              & 5 & 739.3 & 747.5 \\ \hline
Brown  & 0 & 564.6 & 564.6 \\
Corpus & 1 & 407.3 & 408.3 \\
       & 2 & 396.1 & 399.9 \\
       & 3 & 394.9 & 399.4 \\
       & 4 & 394.5 & 399.2 \\
       & 5 & 394.4 & 399.1 \\ \hline
\end{tabular}
\caption{Batch mode prediction.}
\end{center}
\label{batch:tab}
\end{table}

\begin{table}[th]      
\begin{center}  
\begin{tabular}{|l|c|c|} \hline 
& {\small -Log. Likl.} & {\small Posterior Probability} \\ \hline
from god and over wrath grace shall abound   &  74.125 & 0.642 \\ \hline
from god but over wrath grace shall abound   &  82.500 & 0.002 \\
from god and over worth grace shall abound   &  75.250 & 0.295 \\
from god and over wrath grace will abound    &  78.562 & 0.030 \\
before god and over wrath grace shall abound &  83.625 & 0.001 \\
from god and over wrath grace shall a bound  &  78.687 & 0.027 \\
from god and over wrath grape shall abound   &  81.812 & 0.003 \\ \hline
\end{tabular}
\caption{The likelihood induced by a depth four PST for
different corrupted sentences.}
\end{center}
\label{choose:fig}
\end{table}
\begin{table}[th]
\begin{center}
\begin{tabular}{|l|c|c|}
\hline
Model & Test Set ~ (1) & Test Set ~ (2) \\ \hline
Backoff Trigram & 247.7 & 158.7 \\
PST Mixture, $D=2$& 223 & 168 \\
PST Mixture, $D=3$, pruned & 214.1 & 157.6 \\
PST Mixture, $D=3$ & 210.6 & 149.3 \\
\hline
\end{tabular}
\caption{Comparison of different PSTs and a backoff trigram.}
\end{center}
\label{NAB:tab}
\end{table}

Pruning during online adaptation\index{online algorithm!pruning
method} has two advantages. First, it improves memory use. Second,
and less obvious, predictive power may be improved. Rare words tend
to bias the prediction functions at nodes with small counts,
especially if their appearance is restricted to a small portion of
the text. When rare words are removed from the suffix tree, the
estimates of the prediction probabilities at each node are readjusted
to reflect better the probability estimates of the more frequent words.
Hence, part of the bias in the estimation may be overcome.

To support fast insertions, searches and deletions of PST nodes and
word counts we use a hybrid data structure. When we know in advance a
(large) bound on vocabulary size, we represent the root node by arrays
of word counts and possible sons subscripted by word indices.  At
other nodes, we use {\em splay} trees\index{splay tree} \cite{ST85} to
store the branches to longer contexts. Splay trees support search,
insertion and deletion in amortized $O(\log(n))$ time per operation.
Furthermore, they reorganize themselves to decrease the cost of accessing
to the most frequently accessed elements, thus speeding up access to
subtrees associated to more frequent contexts.

\section{Evaluation}
We tested our algorithm in two modes.  In online mode, model structure
and parameters (counts) are updated after each observation.  In batch
mode, the structure and parameters are frozen after the training
phase, making it easier to make comparisons with standard $n$-gram
models. For our smaller experiments we used the Brown corpus, the
Gutenberg Bible, and Milton's Paradise Lost for training and test
material. For comparisons with $n$-gram models, we used ARPA's
North-American Business News\index{ARPA North-American Business (NAB)
News} (NAB) corpus.

For batch training, we partitioned randomly the data into training and
testing sets. We then trained a model by running the online algorithm
on the training set. The resulting model was then frozen and used to
predict the test data.

To illustrate the predictive power of the model, we used it to
generate text by performing random walks over the mixture PST.  A
single step of the random walk was performed by going down the tree
following the current context and stop at a node with the probability
assigned by the algorithm to that node. Once a node is chosen, a word
is picked randomly by the node's prediction function. A result of such
a random walk is given in Figure~\ref{walk:fig}. The PST was trained
on the Brown corpus with maximal depth of five. The output contains
several well formed (meaningless) clauses and also cliches such as
``conserving our rich natural heritage,'' suggesting that the model
captured some longer-term statistical dependencies.

In online mode the advantage of PSTs with large maximal depth is
clear. The perplexity\index{perplexity} of the model
decreases significantly as a function of the depth. Our experiments
so far suggest that the resulting models are fairly insensitive to
the choice of the prior probability, $\alpha$, and a prior which
favors deep trees performed well. Table~1 summarizes the results on
different texts, for trees of growing maximal depth. Note that a
maximal depth $0$ corresponds to a `bag of words' model (zero order),
$1$ to a bigram model, and $2$ to a trigram model.

In our first batch tests we trained the model on 15\% of the data and
tested it on the rest. The results are summarized in Table~2.  The
perplexity obtained in batch mode is clearly higher than that for
online prediction, since only a small portion of the data was used to
train the models. Yet, even in this case the PST of maximal depth
three is significantly better than a full trigram model. In this mode
we also checked the performance of the single MAP model compared to
the mixture of PSTs. This model is found by pruning the tree at the
nodes that obtained the highest confidence value, $L_n(s)$, and using
only the leaves for prediction. As shown in the table, the performance
of the MAP model is consistently worse than the performance of the
mixture.

To illustrate the use of the model in language processing, we applied
it to correcting errors in corrupted text.  This situation models
transcription errors in dictionary-based speech and handwriting
recognition systems, for example. In such systems a language model
selects the most likely alternative between the several options
proposed by previous stages.  Here we used a PST with maximal depth
$4$, trained on $90\%$ of the text of {\em Paradise Lost}. Several
sentences in the held-out test data were corrupted in different ways.
We then used the model in batch mode to evaluate the likelihood of
each of the alternatives.  In Table~3 we demonstrate one such case,
where the first alternative is the correct one. The negative $\log$
likelihood and the posterior probability, assuming that the listed
sentences are all the possible alternatives, are provided.  The
correct sentence gets the highest probability according to the model.

Finally, we trained PST mixtures of varying depths on randomly selected
sentences from the NAB corpus totaling approximately 32.5 million
words and tested it on two corpora: (1) a standard ARPA NAB development test set of around 8 thousand
words, and (2) a separate randomly selected set
of sentences from the NAB corpus, totaling around 2.8 million
words. The results are summarized in Table~4, and compared with a
backoff trigram model \cite{Katz}.

\section{Conclusions and Further Research}
PSTs are able to capture longer correlations than traditional fixed
order $n$-grams, supporting better generalization ability from limited
training data. This is especially noticeable when phrases longer than
a typical $n$-gram order appear repeatedly in the text. The PST
learning algorithm allocates a proper node for the phrase whereas a
bigram or trigram model captures only a truncated version of the
statistical dependencies among words in the phrase.

Our current learning algorithm is able to handle moderate size
corpora, but we hope to adapt it to work with very large training
corpora (100s of millions of words). The main obstacle to those
applications is the space required for the PST. More extensive pruning
may be useful for such large training sets, but the most promising
approach may involve a batch training algorithm that builds a
compressed representation of the final PST from an efficient
representation, such as a suffix array, of the relevant subsequences
of the training corpus.

\end{document}